\documentclass[letterpaper, 10 pt, conference]{ieeeconf}  %

\IEEEoverridecommandlockouts                              %

\usepackage{graphics} %
\usepackage{epsfig} %
\usepackage{times} %
\usepackage{amsmath} %
\usepackage{color}
\usepackage{cite}
\usepackage{hyperref}
\hypersetup{
    colorlinks=true,
    linkcolor=blue,
    filecolor=magenta,      
    urlcolor=cyan,
    pdftitle={Overleaf Example},
    pdfpagemode=FullScreen,
    }
\usepackage{graphicx}
 \usepackage{booktabs} 
 \usepackage[font=footnotesize]{caption}
 \usepackage{amssymb}

\usepackage{booktabs,multirow,pifont,siunitx}
\newcommand{\cmark}{\ding{51}}
\newcommand{\xmark}{\ding{55}}
\usepackage{array}
\usepackage[table]{xcolor}   %
\definecolor{lightgreen}{RGB}{200,255,200}

\newcommand{\dEP}{\Delta_{\text{EP}}}
\newcommand{\tEPone}{\tau_{\text{EP}_1}} %
\newcommand{\tEPtwo}{\tau_{\text{EP}_2}} %
\newcommand{\EPh}{\mathrm{EP}_{\text{H}}}
\newcommand{\EPv}{\mathrm{EP}_{\text{V}}}
\newcommand{\LKh}{\mathrm{LK}_{\text{H}}}
\newcommand{\LKv}{\mathrm{LK}_{\text{V}}}

\title{\LARGE \bf
DriveCritic: Towards Context-Aware, Human-Aligned Evaluation for Autonomous Driving with Vision-Language Models
}

\author{Jingyu Song$\phantom{}^{*1}$\thanks{$^{*}$Work done during an internship at NVIDIA. Corresponding at: \texttt{jingyuso@umich.edu}} \hspace{15pt}
Zhenxin Li$\phantom{}^{2,3}$\hspace{15pt}
Shiyi Lan$\phantom{}^{2}$\hspace{15pt}
Xinglong Sun$\phantom{}^{2}$\hspace{15pt}
Nadine Chang$\phantom{}^{2}$\hspace{15pt} \\
Maying Shen$\phantom{}^{2}$\hspace{15pt}
Joshua Chen$\phantom{}^{2}$\hspace{15pt}
Katherine A. Skinner$\phantom{}^{1}$\hspace{15pt}
Jose M. Alvarez$\phantom{}^{2}$\vspace{8pt}\\
$\phantom{}^1$University of Michigan\hspace{5pt}
$\phantom{}^2$NVIDIA\hspace{5pt}
$\phantom{}^3$Fudan University
}

\renewcommand{\baselinestretch}{0.98}

\begin{document}

\maketitle
\thispagestyle{empty}
\pagestyle{empty}

\begin{abstract}
Benchmarking autonomous driving planners to align with human judgment remains a critical challenge, as state-of-the-art metrics like the Extended Predictive Driver Model Score (EPDMS) lack context awareness in nuanced scenarios. To address this, we introduce DriveCritic, a novel framework featuring two key contributions: the DriveCritic dataset, a curated collection of challenging scenarios where context is critical for correct judgment and annotated with pairwise human preferences, and the DriveCritic model, a Vision-Language Model (VLM) based evaluator. Fine-tuned using a two-stage supervised and reinforcement learning pipeline, the DriveCritic model learns to adjudicate between trajectory pairs by integrating visual and symbolic context. Experiments show DriveCritic significantly outperforms existing metrics and baselines in matching human preferences and demonstrates strong context awareness. Overall, our work provides a more reliable, human-aligned foundation to evaluating autonomous driving systems. The project page for DriveCritic is \url{https://song-jingyu.github.io/DriveCritic}.

\end{abstract}

\section{Introduction}

Planning is one of the central components to enable autonomous driving, as it is expected to predict safe and efficient future trajectories for the autonomous vehicle to follow~\cite{chen2024e2e_survey, mllm_driving_survey}. Recently, end-to-end (E2E) driving systems that are trained with planning-oriented goals have advanced at a fast pace and demonstrated superior performance in large-scale benchmarks~\cite{caesar2021nuplan,li2024hydramdp, li2025hydranext, chen2024vadv2, caesar2020nuscenes}. However, benchmarking planners in a way that accurately reflects safety and human expectations still remains challenging~\cite{Dauner2024NEURIPS, zhai2023ad-mlp}. Without this property, a driving planner can achieve state-of-the-art (SOTA) performance on standard quantitative metrics, yet remain misaligned with actual human preferences on nuanced scenes in real driving scenarios.

Evaluation of driving policies is typically categorized into two main approaches: closed-loop simulation, employed by platforms like CARLA \cite{dosovitskiy2017carla}, offers high-fidelity, interactive testing where the agent's actions influence the subsequent states of the environment. While considered the gold standard for assessing real-world performance, it is computationally expensive, suffers from a simulation-to-reality gap, and is difficult to scale for comprehensive testing across diverse scenarios \cite{Dauner2024NEURIPS, chen2024e2e_survey}. In contrast, open-loop evaluation replays logged sensor data from real-world driving scenes and assesses the planner's predicted trajectory without affecting the behavior of other agents. This approach is highly scalable, data-driven, and allows for direct comparison on massive, real-world datasets, making it the preferred method for large-scale benchmarks~\cite{caesar2020nuscenes, waymo_womd_2021, sun2020waymo_open_dataset, caesar2021nuplan}. While closed-loop testing better reflects human driving preferences than open-loop simulation~\cite{chen2024e2e_survey}, our work takes a different path: rather than first aligning open-loop evaluation with closed-loop performance, we propose a solution towards directly bridging open-loop evaluation to expert human alignment.

\begin{figure}[t]
    \centering
    \includegraphics[width=0.97\linewidth]{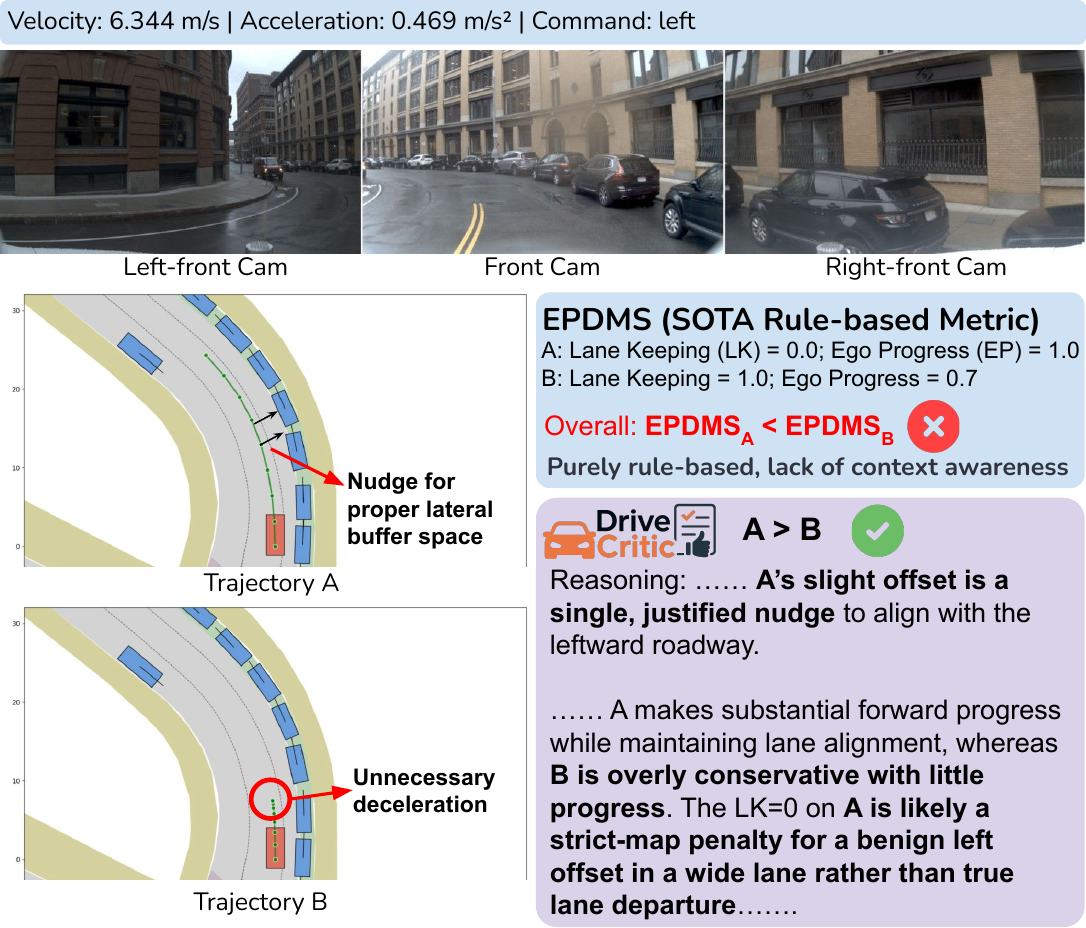}
    \caption{Example from NAVSIM \cite{Dauner2024NEURIPS} illustrating the need for context-aware evaluation in autonomous driving. In this pairwise comparison task, trajectory A briefly nudges left to maintain a safe lateral buffer - an action that is contextually appropriate and not a true lane departure. Our DriveCritic model not only prefers A but also generates similar reasoning, demonstrating its contextual understanding capability. By contrast, the SOTA rule-based metric EPDMS~\cite{Cao2025CORL} assigns a lower score to A and favors B simply because B remains within a fixed lane-keeping threshold despite its unnecessarily low progress. Key BEV legend: Ego vehicle - red rectangle at the center (0 m, 0 m) oriented upward; Trajectory waypoints - green dotted line with circular points (8 waypoints for a 4 s horizon, spaced 0.5 s apart) starting at the ego vehicle’s rear-axle center. Best viewed zoomed in and in color.}
    \label{fig:pitch}
    \vspace{-7mm}
\end{figure}

To understand the necessity of the proposed approach, it is crucial to first examine the limitations of prior methods within the open-loop paradigm. Early evaluation methods predominantly relied on simple displacement errors like Average Displacement Error (ADE) and Final Displacement Error (FDE) \cite{caesar2020nuscenes, zhai2023ad-mlp}. However, these metrics are insufficient for the multimodal and safety-critical nature of driving~\cite{Dauner2024NEURIPS, zhai2023ad-mlp}, as they often penalize valid alternative driving behaviors by constraining the notion of correctness to a single reference trajectory, and fail to capture crucial aspects like collision avoidance or passenger comfort.

A more recent proposal, the Rater Feedback Score (RFS)~\cite{waymo2025E2Echallenge}, attempts to tackle these limitations by relying on expert annotators who provide three reference trajectories with different scores; each candidate is then scored based on the closest rater-specified trajectory. While RFS better accommodate multimodality than ADE/FDE, it suffers from limited scalability due to costly human annotation and lacks interpretability of its scores (e.g., the meaning of a 7 versus an 8 remains unclear) because of the absence of a public scoring rubric.

To address the shortcomings of imitation-based metrics that rely on displacement errors, state-of-the-art benchmarks have introduced more comprehensive, rule-based metrics. A prominent example is the Predictive Driver Model Score (PDMS), and its Extended version EPDMS, proposed with the NAVSIM benchmark~\cite{Dauner2024NEURIPS}. EPDMS is a rule-based evaluator that considers critical factors such as safety, comfort, and progress, and has been widely adopted for evaluating modern driving policies~\cite{Cao2025CORL, li2025gtrs, styledrive2025, li2025recogdrive}.

Despite the widespread adoption of the SOTA rule-based metrics like PDMS/EPDMS, we observe and argue that they still suffer from a fundamental limitation: a lack of context awareness and human alignment (Fig.~\ref{fig:pitch}). We define \emph{human alignment} as the ability to evaluate driving plans in a way that reflects how experienced human drivers balance safety, progress, and social norms in complex traffic situations. Specifically, these metrics operate on a predefined set of fixed rules and thresholds, which struggle to capture such human-like judgment in these situations.
For instance, a minor lane deviation to create a safe lateral buffer space might be heavily penalized, or an overly aggressive trajectory that ignores the stop sign might be scored favorably. We reveal this deficiency by analyzing EPDMS scores on expert human trajectories in challenging scenarios and curating a pairwise preference dataset that concentrates on such ambiguous regimes, where EPDMS often diverges from human preferences.

To bridge this critical gap, we propose DriveCritic, a novel framework towards human-aligned evaluation of autonomous driving planners.
We first introduce the DriveCritic dataset, a piloting collection of challenging and ambiguous driving scenarios where existing metrics often fail, annotated with pairwise human preferences.
Second, we introduce the DriveCritic model as an “expert-human-aligned” judge, which leverages powerful contextual reasoning and common-sense knowledge of Vision-Language Models (VLMs) \cite{achiam2023gpt4_report, openai2025gpt5, openai2025o3, team2023gemini, Qwen2.5-VL}. By fine-tuning a VLM through reinforcement learning from verifiable rewards (RLVR) paradigm~\cite{shao2024deepseekmath, yu2025dapo, yue2025rlvr}, 
DriveCritic achieves SOTA alignment with expert human preferences, setting a reliable foundation towards developing context-aware, human-aligned evaluation for autonomous driving.
Our main contributions can be summarized as follows:
\begin{itemize}
\item We identify and demonstrate the limitations of state-of-the-art rule-based metrics like EPDMS, showing their lack of context awareness and alignment with expert human judgment in nuanced driving scenarios.
\item We introduce the DriveCritic dataset, a curated dataset sampled from NAVSIM~\cite{Dauner2024NEURIPS} for assessing driving evaluation methods, featuring challenging scenarios annotated with pairwise expert human preferences.
\item We propose the DriveCritic model, a novel VLM-based model that is fine-tuned with the RLVR pipeline to evaluate driving trajectories, and show that it significantly outperforms existing metrics in aligning with human expert preferences, achieving 76\% accuracy on the proposed DriveCritic dataset.
\end{itemize}

\section{Related Works}
\subsection{Benchmarking Autonomous Driving}
\label{sec:related_works_av_benchmarks}

Evaluating the performance of autonomous driving systems is a complex and multifaceted challenge. Current methodologies are largely split between two paradigms: \emph{closed-loop simulation} and \emph{open-loop evaluation}.

Closed-loop simulation~\cite{caesar2021nuplan, dosovitskiy2017carla, jia2024bench2drive} places the autonomous agent in an interactive, simulated environment where its actions directly influence future states.
While often considered the gold standard for benchmarking autonomous driving due to its interactive nature, this approach is computationally intensive, struggles to scale to the diversity of real-world scenarios, and can struggle with the persistent sim-to-real domain gap~\cite{Dauner2024NEURIPS, Cao2025CORL}.

On the other hand, open-loop evaluation~\cite{hu2023uniad, caesar2020nuscenes} leverages real-world log-replays and human trajectories for benchmarking, offering scalability and interpretability~\cite{chen2024e2e_survey}. Early open-loop evaluation methods (e.g., ADE and FDE) rely heavily on comparing the displacement error from the human trajectory. These methods are simple to compute but they usually fail to capture the multimodal nature of driving or to penalize unsafe behavior~\cite{jia2024bench2drive, zhai2023ad-mlp, Dauner2024NEURIPS, ego_status_all_you_need}. To address these shortcomings, recent benchmarks have proposed rule-based scoring systems that explicitly evaluate safety compliance, progress, and comfort instead of simply focusing on displacement errors, encouraging the multimodal nature of driving while ensuring safety~\cite{caesar2021nuplan, Dauner2024NEURIPS}. NAVSIM~\cite{Dauner2024NEURIPS} and its successor Pseudo-Simulation~\cite{Cao2025CORL} advance this paradigm by simulating trajectories in a symbolic space of objects and maps and scoring them with the EPDMS metric, a comprehensive rule-based suite. While EPDMS has become the state-of-the-art for scalable open-loop evaluation, its logic remains hard-coded and limited to symbolic representations, which makes it inherently ``context-blind,'' as it lacks access to the rich visual and semantic cues that a human driver uses to navigate socially complex or ambiguous situations~\cite{styledrive2025,mu2024most}. Our work directly addresses this gap by proposing a VLM-based evaluator that can complement these rule-based metrics with context-aware and human-like reasoning.

\subsection{VLMs in Autonomous Driving}
Recent advances in Large Language Models (LLMs) and VLMs~\cite{achiam2023gpt4_report, team2023gemini} have motivated a wave of research into their application for autonomous driving across a wide range of tasks~\cite{tiandrivevlm, sima2024drivelm, wang2025omnidrive, jiang2024senna, jiang2025alphadrive, qian2025agentthink}. While these methods demonstrate the strong potential of VLMs for driving scene understanding and decision making, we note that leveraging VLMs for driving evaluation remains less explored.

Motivated by the progress in using LLMs/VLMs as a judge in other domains~\cite{ye2024justice, li2025generation, liualigning}, researchers in~\cite{you2025comprehensive, xie2025drivebench} have started to explore using VLMs as evaluators of driving behaviors. Furthermore, HE-Drive~\cite{wang2024he} proposes to incorporate a VLM-guided scorer to help adjust driving styles while ensuring comfort. Meanwhile, a closely related work, StyleDrive~\cite{styledrive2025}, leverages a fine-tuned VLM to mine scenarios of different driving styles and develops a style-aware metric by adjusting key sub-metrics in EPDMS~\cite{Cao2025CORL} according to the annotated driving styles. Our work, DriveCritic, shares the same motivation as StyleDrive~\cite{styledrive2025} on improving the context awareness of EPDMS while making a distinct contribution on VLM usage and task formulation. We conduct a systematic study on misalignment between EPDMS and expert human preferences, and position a VLM-based model as a context-aware evaluator capable of generating human-aligned pairwise judgment on ambiguous scenarios. Notably, DriveCritic can be seamlessly integrated into frameworks like TrajHF~\cite{li2025trajhf} by supplying scalable, human-aligned preference signals to guide trajectory generation under its reinforcement-learning-from-human-feedback pipeline. Moreover, the DriveCritic model is fine-tuned using the RLVR paradigm~\cite{shao2024deepseekmath, yu2025dapo, yue2025rlvr} following the success of pioneering works in autonomous driving~\cite{jiang2025alphadrive, qian2025agentthink, li2025recogdrive}.

\begin{table}[t]  %
  \centering
  \caption{
    EPDMS and sub-scores of human expert trajectories on the
    \emph{navtrain} and \emph{navtest} splits of NAVSIM~\cite{Dauner2024NEURIPS}.
    Abbreviations in Sec.~\ref{sec:epdms}.
  }
  \label{tab:human_epdms_scores_abbrev}
  \resizebox{0.49\textwidth}{!}{%
  \begin{tabular}{@{}lcccccccccc@{}}
    \toprule
    \textbf{Split} & \textbf{NC} & \textbf{DAC} & \textbf{DDC} & \textbf{TLC}
    & \textbf{EP} & \textbf{TTC} & \textbf{LK} & \textbf{HC} & \textbf{EC} & \textbf{EPDMS}\\
    \midrule
    \emph{navtrain} & 1.00 & 1.00 & 1.00 & 0.98 & \textbf{0.88} & 1.00 & \textbf{0.90} & 0.98 & 0.91 & 0.92 \\
    \emph{navtest}  & 1.00 & 1.00 & 1.00 & 0.97 & \textbf{0.87} & 1.00 & \textbf{0.87} & 0.98 & 0.90 & 0.90 \\
    \bottomrule
  \end{tabular}
  } %
  \vspace{-6mm}
\end{table}

\section{Preliminaries}
\label{sec:preliminaries}
This work's focus on addressing the gap of context awareness in the rule-based evaluation method,
EPDMS~\cite{Cao2025CORL}, is grounded in its status as the SOTA open-loop metric, which has been discussed in Sec.~\ref{sec:related_works_av_benchmarks}. We begin by reviewing the technical details of EPDMS based on the NAVSIM benchmark~\cite{Dauner2024NEURIPS, Cao2025CORL}, and then we discuss its limitations that motivate our work.

\subsection{EPDMS}
\label{sec:epdms}
EPDMS is a comprehensive rule-based metric proposed with the NAVSIM benchmark~\cite{Cao2025CORL} that focuses on challenging scenarios in the OpenScene dataset~\cite{contributors2023openscene}, a lightweight redistribution of nuPlan~\cite{caesar2021nuplan}. It evaluates a fixed-horizon trajectory (typically 4\,s) using ground-truth perception (e.g., object bounding boxes, BEV maps) with an interpretable set of rule-based sub-metrics capturing safety compliance, progress, and comfort. In practice, it combines \emph{multiplicative} penalties for safety rule violations with a \emph{weighted average} of trajectory-quality sub-scores:
\[
\mathrm{EPDMS} =
\left(\prod_{m\in\mathcal{M}_{\mathrm{pen}}} s_m\right)\cdot
\frac{\sum_{m\in\mathcal{M}_{\mathrm{avg}}} w_m s_m}{\sum_{m\in\mathcal{M}_{\mathrm{avg}}} w_m},
\]
where $\mathcal{M}_{\mathrm{pen}} =$ {No at-fault Collisions (NC), Drivable Area Compliance (DAC), Driving Direction Compliance (DDC), Traffic Light Compliance (TLC)}
and $\mathcal{M}_{\mathrm{avg}} =$ {Time to Collision (TTC), Ego Progress (EP), Lane Keeping (LK), History Comfort (HC), Extended Comfort (EC)}.
Here, $s_m \in [0,1]$ denotes the sub-scores for metric $m$: a value of $1$ indicates full compliance, $0$ indicates a hard violation, and fractional values (e.g., $0.7$ for EP) capture partial compliance depending on the rule. 
The term $w_m$ denotes the relative weight assigned to each averaged sub-metric, reflecting their importance in EPDMS.
The full specification of EPDMS can be found in~\cite{Cao2025CORL}.

\subsection{Context Gap}
\label{sec:limitation_epdms}
As noted in~\cite{caesar2021nuplan, Dauner2024NEURIPS}, the human driver trajectories in NAVSIM can be considered as expert demonstrations driven by trained operators. This raises a natural consistency check: if EPDMS is truly aligned with expert human preferences, the human trajectories would be expected to achieve perfect scores. However, as shown in Table~\ref{tab:human_epdms_scores_abbrev}, this is not the case. While safety-critical sub-scores such as NC, DAC, DDC, TLC, and TTC saturate near 1.0 for human driving, two sub-scores consistently fall behind: Ego Progress (EP) and Lane Keeping (LK). While Extended Comfort (EC) is also lower than the aggregated EPDMS, we do not analyze it further because comfort experience is inherently subjective and less reliably assessed from visual inspection. For clarity, we now detail the computation of LK and EP, as these sub-scores play a central role in our analysis.

\begin{figure*}[t]
    \centering
    \includegraphics[width=1.0\linewidth]{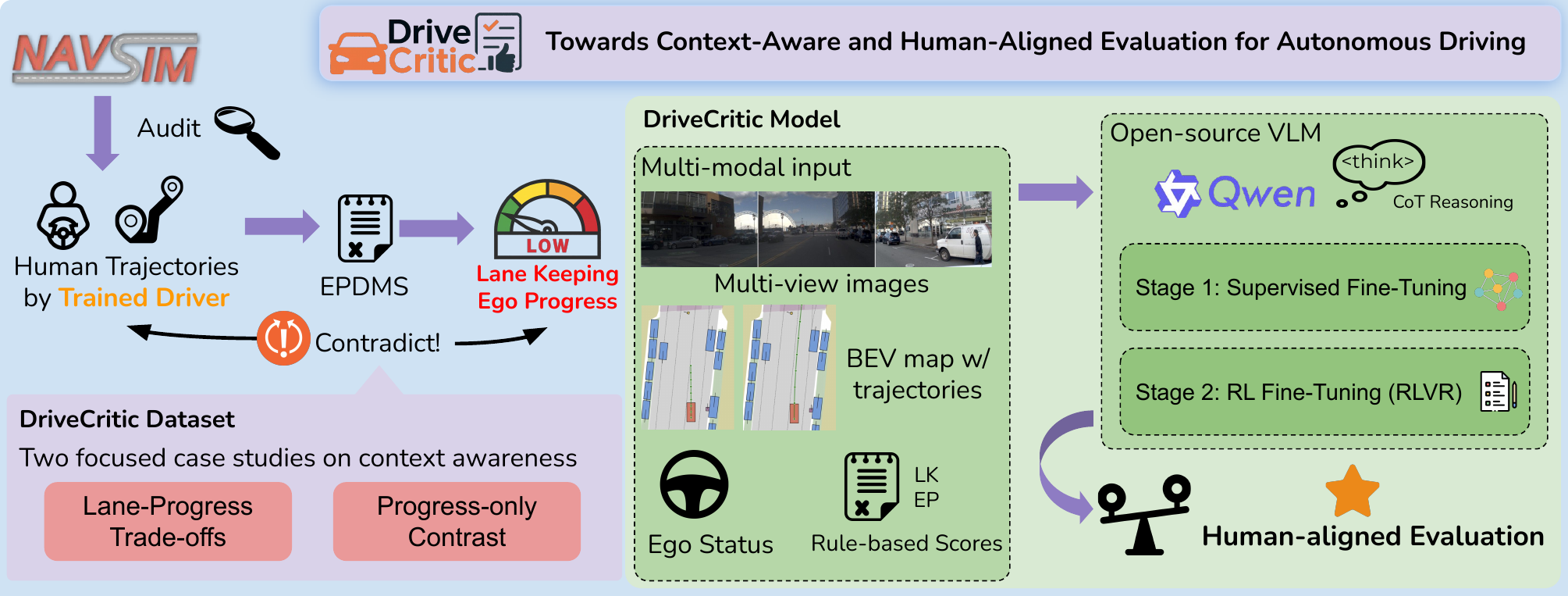}
    \caption{An overview of the DriveCritic framework. The DriveCritic dataset is a pilot benchmark focusing on context-aware evaluation. The DriveCritic model integrates rich multi-modal inputs and is fine-tuned with a two-stage training procedure, enabling it to generate human-aligned evaluation decisions in challenging driving scenarios.}
    \label{fig:overview}
    \vspace{-5mm}
\end{figure*}

\noindent \textbf{LK} checks whether the ego vehicle stays within its lane without prolonged deviation. At each simulation step the lateral offset from the lane center is measured; a violation occurs only if $d>0.5$ m for more than 2 s. The final score is binary (1 if no sustained violation, else 0).

\noindent \textbf{EP} measures route advancement relative to a context-blind upper bound $d_\mathrm{ref}$ from the Predictive Driver Model (PDM)-Closed planner.~\cite{Dauner2024NEURIPS}:
\[
\mathrm{EP}=\min\!\left(1,\frac{d_\mathrm{ego}}{d_\mathrm{ref}}\right),
\]
with scores clipped to $[0,1]$. If $d_{\text{ref}} < 5\,\text{m}$, the ratio is discarded to avoid unstable cases.

Auditing low-score scenes reveals that human experts often make context-appropriate lane nudges or reduce progress to accommodate conservative cues (examples in Fig.~\ref{fig:pitch} and Fig.~\ref{fig:qualitative}). Because EPDMS penalizes these desirable behaviors, we use LK and EP as probes to mine such nuanced cases and build our evaluation dataset.

\section{Technical Approach}
In this section, we present the \emph{DriveCritic} framework, covering both the DriveCritic dataset construction and the DriveCritic model design.
\subsection{DriveCritic Dataset}
The DriveCritic dataset is sampled and constructed from NAVSIM~\cite{Dauner2024NEURIPS}, comprising 5,730 trajectory pairs curated as a pilot benchmark to highlight the need for context-aware evaluation. The construction process is detailed below.

\subsubsection{Dataset Construction Strategy}

As discussed in Sec.~\ref{sec:limitation_epdms}, we extend our audit of EPDMS of human trajectories and mine ambiguous scenarios from NAVSIM~\cite{Dauner2024NEURIPS} through the lane keeping (LK) and ego progress (EP) scores (Fig.~\ref{fig:overview}). However, quantitatively reducing human preferences to a single numeric score for a trajectory is challenging, as no widely accepted rubric exists for grading nuanced trade-offs (e.g., minor lane offsets to bypass a stopped vehicle). We therefore formulate the dataset task as a pairwise adjudication problem~\cite{liualigning, qin2024large} and augment the human trajectories with samples from a large static vocabulary~\cite{li2025gtrs, chen2024vadv2}, paired with their raw EPDMS sub-scores.

\begin{table}[t]
  \centering
  \caption{Trajectory sampling pattern for the two focused case studies. Each sampled trajectory pair consists of the human trajectory and a vocabulary trajectory that matches the sub-scores pattern. The other sub-scores of the sampled trajectories are perfect.}
  \label{tab:case_rules}
  \resizebox{0.48\textwidth}{!}{%
  \begin{tabular}{@{}lcc@{}}
    \toprule
    \textbf{Case} & \textbf{Human (H)} & \textbf{Vocabulary (V)} \\
    \midrule
    \multirow{2}{*}{1}
      & $\LKh = 0,\; \EPh \ge \tEPone$ 
      & $\LKv = 1,\; \EPv \le \EPh - \dEP$ \\
      & \cellcolor{gray!15}$\LKh = 1,\; \EPh \le \tEPtwo$
      & \cellcolor{gray!15}$\LKv = 0,\; \EPv \ge \EPh + \dEP$ \\
    2
      & $\EPh \le \tEPtwo$ 
      & $\EPv \ge \EPh + \dEP$ \\
    \bottomrule
  \end{tabular}%
  }
\par\vspace{3pt}   %
\raggedright\footnotesize
  \textbf{Hyperparameters:} $\tEPone=0.88$, $\tEPtwo=0.75$, $\dEP=0.2$.
  \vspace{-2mm}
\end{table}

\begin{table}[t]
\centering
\caption{Number of trajectory pairs by split and data source. 
Case~1: lane-progress trade-off; Case~2: progress-only contrast.}
\label{tab:dc_dataset_stats}
\begin{tabular}{llrrr}
\toprule
\textbf{Split} & \textbf{Data Source} & \textbf{Case~1} & \textbf{Case~2} & \textbf{Total} \\
\midrule
Train & \emph{navtrain} & 2626 & 1938 & 4564 \\
Test  & \emph{navtest}  & 663  & 503  & 1166 \\
\midrule
Total & --               & 3289 & 2441 & 5730 \\
\bottomrule
\end{tabular}
\vspace{-6mm}
\end{table}

In the process of inspecting scenarios where human trajectories receive low LK or EP score, we observe that EPDMS usually misjudges two characteristic scene types. First, in scenarios where human drivers receive $\text{LK}=0$, human drivers may briefly sacrifice lane keeping to maintain progress in scenarios such as deviating slightly to bypass a stopped vehicle. However, vocabulary trajectories that strictly remain in-lane ($\text{LK}=1$), even with noticeably lower progress, are often scored more favorably by EPDMS.
Second, in scenarios where a human driver receives a lower than typical $\text{EP}$ score, human driving frequently reflects justifiable conservative driving behaviors where reduced progress is contextually appropriate, while vocabulary trajectories with notably higher EP are not always preferable, as overly aggressive progress can conflict with safe or courteous driving. Consequently, we form two diagnostic case studies constructed following the rules in Table~\ref{tab:case_rules}:

    \noindent \textbf{Case~1 (Lane-Progress Trade-off):} We first sample pairs where the \emph{human} has $\text{LK}=0$ and high EP versus a \emph{vocabulary} alternative with $\text{LK}=1$ and lower EP (first row in Table~\ref{tab:case_rules}); in practice, we find that the human trajectory is preferred in the majority of such pairs after annotation. To prevent a degenerate rule (``always choose $\text{LK}=0$'') learned by the evaluators, we additionally include \emph{mirror} pairs (second row in Table~\ref{tab:case_rules}). For the mirror pairs, we sample a vocabulary candidate that has $\text{LK}=0$ and high EP while the human has $\text{LK}=1$ with lower EP. This forces the model to reason about context rather than keying on LK alone.
    
  \noindent \textbf{Case~2 (Progress-only Contrast):}  
    In this case, we sample human trajectories with $\text{EP}$ lower than a pre-defined threshold, with its paired vocabulary trajectory receiving a notably higher $\text{EP}$ while other sub-scores of both trajectories are perfect (third row in Table~\ref{tab:case_rules}). These pairs focus on context-aware evaluation on the driving progress.

We list the hyper-parameters used in Table~\ref{tab:case_rules}, where the $\text{EP}$ thresholds $\tEPone$ and $\tEPtwo$ are set empirically from NAVSIM statistics, and the progress margin $\dEP$ ensures a visually clear separation in $\text{EP}$.
Together, these carefully constructed cases form the backbone of the DriveCritic dataset, providing controlled yet diverse scenarios where rule-based EPDMS scoring and human driving preferences often diverge.

\subsubsection{Human Preferences Annotation}
\label{sec:technical_approach_annotation}
After sampling, each trajectory pair is randomly assigned as A or B for human preferences annotation.  
Table~\ref{tab:dc_dataset_stats} reports the resulting dataset size.
We create train/test splits with verified labels on the test set and scalable auto-labels on the train set (details below). 

We recruit the main author to annotate the entire test split, ensuring consistent labeling criteria. The annotator can be regarded as a domain expert, with over five years of research experience in autonomous driving, thereby providing reliable ground-truth preferences. During the annotation process, a guideline was iteratively refined, and a subsequent verification process was conducted to ensure that all labels adhered to this guideline. We also discard samples that are too ambiguous to judge, ensuring that each retained pair exhibits a clear and discernible preference.

On the test set, we observe that preferences are highly skewed in the lane–progress trade-off (Case~1): human trajectories are chosen in $608/663$ pairs ($91.7\%$), whereas preferences are more balanced in the progress-only contrast (Case~2), with humans chosen in $304/503$ pairs ($60.4\%$). This indicates that Case~1 is comparatively unambiguous: humans are almost always preferred, whether they briefly nudge out of lane to maintain progress or remain more conservative to preserve lane keeping. In contrast, Case~2 reflects genuine ambiguity, where conservative progress is sometimes favored and sometimes penalized. To scale annotation for the \emph{train} split, we therefore use \emph{pseudo-labels} (human-preferred) for Case~1 and employ GPT-5~\cite{openai2025gpt5} to annotate Case~2. When prompted with specific instructions distilled from the annotation guideline of Case~2, GPT-5 achieves high accuracy ($82\%$) on the verified test set. The exact prompt will be made available on our project website.

\subsection{DriveCritic Model}

\subsubsection{Model Design}
The goal of the DriveCritic model is to adjudicate between candidate trajectory pairs in challenging driving scenarios, producing pairwise judgments that align with human preferences. Motivated by the significant success of integrating and fine-tuning VLMs in autonomous driving tasks such as perception, reasoning, and planning~\cite{jiang2025alphadrive, tiandrivevlm, li2025recogdrive, qian2025agentthink}, we also leverage an open-source VLM~\cite{Qwen2.5-VL} as the backbone of the DriveCritic model. Specifically, we adopt the 7B variant of the Qwen2.5-VL model family~\cite{Qwen2.5-VL}, as it provides a favorable balance between training efficiency and reasoning capability. As shown in Fig.~\ref{fig:overview}, the DriveCritic model conditions the VLM on four inputs: (i) a stitched three-camera view (left-front, front, right-front) following~\cite{wang2025omnidrive, qian2025agentthink}, (ii) a BEV map with scene context (e.g., drivable area, lanes, crosswalks, nearby agents) where the two candidate trajectories are overlaid separately to avoid overlap, (iii) the ego-vehicle status (i.e., current acceleration, velocity, driving command), and (iv) the EPDMS sub-scores Ego Progress (EP) and Lane Keeping (LK). We experimented with alternative configurations such as feeding raw waypoint coordinates in the text prompt, including additional EPDMS sub-scores, or projecting candidate trajectories onto the camera view, but empirically found that the chosen setup has the most reliable performance.

The VLM is prompted as an expert driving evaluator, tasked with selecting the more reasonable trajectory between A and B (i.e., the two candidate trajectories of each scenario in the DriveCritic dataset). The prompt specifies role, inputs, and evaluation scope (with emphasis on EP and LK given current context), and follows~\cite{shao2024deepseekmath, yu2025dapo} to enforce a structured reasoning process followed by a single preference decision. This design guides the model to cross-check visual and symbolic cues, understanding and reasoning about the appropriateness of LK and EP sub-scores, and yielding human-aligned pairwise judgments.

\subsubsection{Two-stage Training Pipeline}
\label{sec:two_stage_training}
Our initial attempts with reinforcement learning (RL) alone proved unstable, with the model requiring a long warm-up before showing meaningful improvement. To address this, we adopt a two-stage pipeline of \emph{supervised fine-tuning (SFT)} followed by \emph{RL fine-tuning}:

\noindent \textbf{Supervised Fine-Tuning:}
We first fine-tune the base Qwen2.5-VL-7B model on a subset of 1,100 pairs randomly sampled from the training split of the DriveCritic dataset. For each pair, we employ GPT-5~\cite{openai2025gpt5} as a ``teacher'' model. For each trajectory pair, the teacher model is prompted with the ground-truth human preferences label and tasked with generating a corresponding chain-of-thought reasoning trace. This stage helps to warm up the model’s ability to follow the required response format and to ground its judgments in step-by-step reasoning before RL.

\noindent \textbf{Reinforcement Learning Fine-Tuning:} 
In the second stage, we refine the model from the SFT stage using the RLVR paradigm. Specifically, we adopt the Decoupled Clip and Dynamic Sampling Policy Optimization (DAPO) algorithm~\cite{yu2025dapo}, a recent advancement built upon Group Relative Policy Optimization (GRPO)~\cite{shao2024deepseekmath} that improves training efficiency and stability. Like GRPO, DAPO avoids the need for an explicit value function by computing relative advantages within a group of samples, while further introducing mechanisms that stabilize updates and accelerate convergence. We use the same reward design as~\cite{shao2024deepseekmath}, encouraging both format adherence (e.g., correct use of the \texttt{<think>} token) and accuracy, as the original reward design in DAPO was found to introduce training instability in our setting. Due to space limitations, we do not include full algorithmic details here and instead refer readers to the original GRPO and DAPO papers for comprehensive descriptions~\cite{shao2024deepseekmath, yu2025dapo}.

\section{Experiments and Results}

\subsection{Implementation Details}
We summarize the key implementation details of the DriveCritic model. As described in Sec.~\ref{sec:two_stage_training}, training proceeds in two stages.
In the first stage, supervised fine-tuning (SFT) is performed on 1,100 annotated trajectory pairs with reference reasoning traces generated by GPT-5~\cite{openai2025gpt5}. We fine-tune for 5~epochs with a per-device batch size of~1 and a learning rate of $1\times10^{-4}$ using the LoRA (Low-Rank Adaptation) method implemented in LLaMA-Factory~\cite{zheng2024llamafactory}. In the second stage, RL fine-tuning is applied under the RLVR paradigm using the train set of the DriveCritic dataset. We adopt the EasyR1 library~\cite{zheng2025easyr1} built on the verl framework~\cite{sheng2024verl}, training for 4~epochs using bfloat16 data type on 16~NVIDIA A100 GPUs with a global batch size of~256, a rollout number of~8, and a learning rate of $1\times10^{-6}$. The rollout temperature is set to~1.0 to encourage exploration, while validation is performed with a temperature of~0.1 for stable evaluation. The same training configuration is used for all model variants reported in the ablation studies (Sec.~\ref{sec:ablations}).

\begin{figure*}[t]
    \centering
    \includegraphics[width=1.0\linewidth]{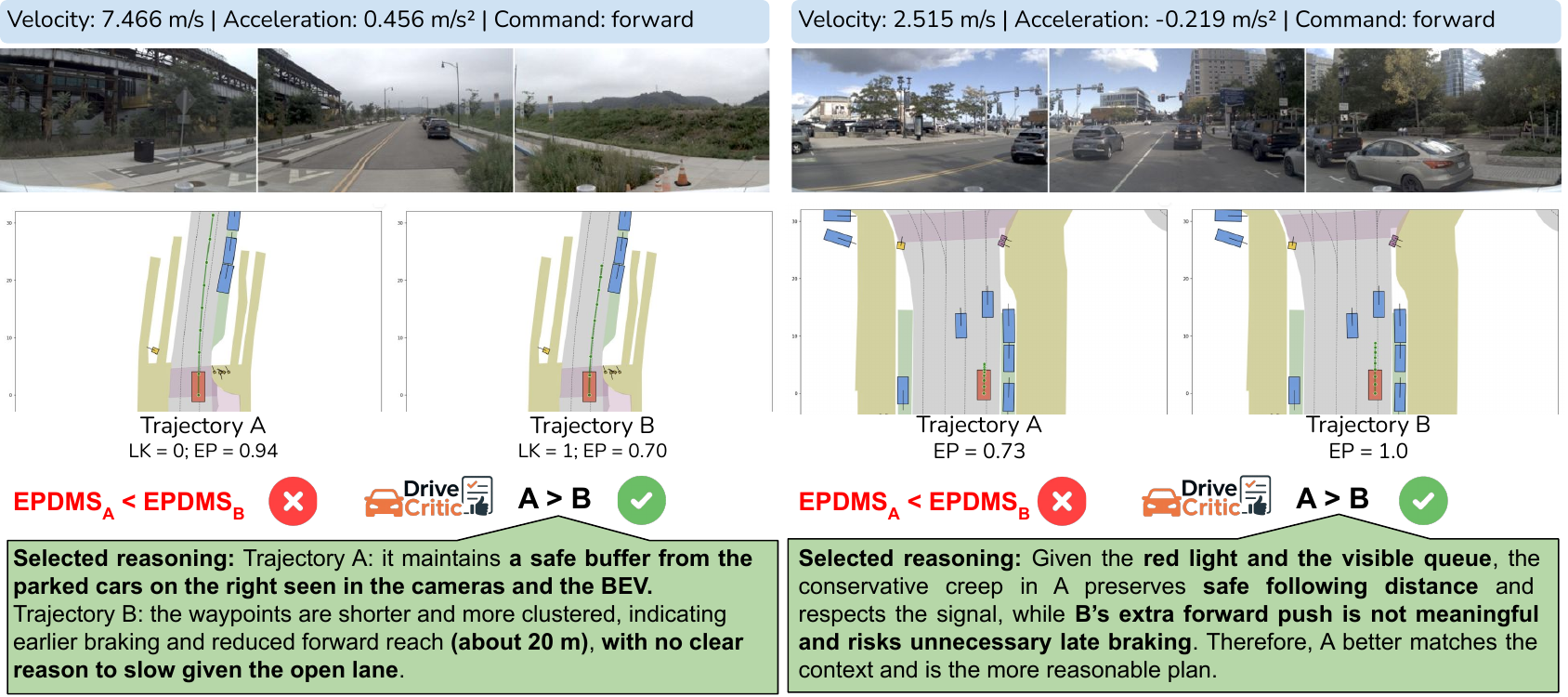}
    \caption{Qualitative examples illustrating DriveCritic’s contextual understanding and evaluation ability. Only representative reasoning steps are shown due to space constraints. Key BEV legend: Ego vehicle - red rectangle at the center (0 m, 0 m) oriented upward; Trajectory waypoints - green dotted line with circular points (8 waypoints for a 4 s horizon, spaced 0.5 s apart) starting at the ego vehicle’s rear-axle center. Best viewed zoomed in and in color.}
    \label{fig:qualitative}
    \vspace{-6mm}
\end{figure*}

\begin{table}[t]
\centering
\caption{Overall accuracy on the DriveCritic test set. ``Fine-tuning'' indicates whether the model was fine-tuned on DriveCritic data beyond its original pretraining.
}
\label{tab:overall_comparison}
\begin{tabular}{lcc}
\toprule
\textbf{Method} & \textbf{Fine-tuning} & \textbf{Accuracy} \\
\midrule
EPDMS~\cite{Cao2025CORL} & \xmark & 0.414 \\
OpenAI-o3 (zero-shot)~\cite{openai2025o3} & \xmark & 0.533 \\
GPT-5 (zero-shot)~\cite{openai2025gpt5} & \xmark & 0.552 \\
Qwen2.5-VL-7B (zero-shot) & \xmark & 0.480 \\
Supervised Pairwise Classifier & \cmark & 0.648 \\
\textbf{DriveCritic (ours)} & \cmark & \textbf{0.760} \\
\bottomrule
\end{tabular}
\vspace{-6mm}
\end{table}

\subsection{Overall Comparison}
\label{sec:overall-comparison}
We first evaluate all methods on the DriveCritic dataset, using the verified test split as described in Sec.~\ref{sec:technical_approach_annotation}. The primary evaluation metric is \emph{accuracy}, defined as the proportion of pairwise comparisons in which the model’s judgment agrees with the human-preferred trajectory.

\subsubsection{Baselines}
We compare DriveCritic against a wide range of baselines covering rule-based metrics, general-purpose LLMs, and controlled supervised models:  

\noindent \textbf{Rule-based:} EPDMS~\cite{Cao2025CORL} serves as the SOTA rule-based benchmark. Since EPDMS outputs a scalar score per trajectory, we select the higher-scoring trajectory as its preference.

\noindent \textbf{General VLMs:} We evaluate SOTA closed-source (OpenAI-o3~\cite{openai2025o3} and GPT-5~\cite{openai2025gpt5}) and open-source (Qwen2.5-VL-7B~\cite{Qwen2.5-VL}) VLMs under the same evaluation prompt as DriveCritic. This baseline captures the out-of-the-box reasoning ability of frontier VLMs without domain-specific fine-tuning.

\noindent \textbf{Supervised Pairwise Classifier:} We implement a supervised pairwise classifier as a data-driven baseline that does not rely on VLMs. The model employs ResNet-101~\cite{he2016resnet} encoders for stitched camera images and BEV maps with overlaid candidate trajectories, concatenated with feature encodings of the ego status and EPDMS sub-scores through an MLP-based fusion layer. The classifier is trained on the \emph{train} split of the DriveCritic dataset with cross-entropy loss for 20 epochs, and results are reported from the best checkpoint. This baseline provides a learning-based alternative, highlighting the benefits of a VLM backbone in DriveCritic.

\subsubsection{Results}
Table~\ref{tab:overall_comparison} reports the overall accuracy of all baselines and DriveCritic on the DriveCritic test set. The rule-based EPDMS metric performs the weakest, reflecting the pressing need to improve the context awareness in rule-based driving metrics. General-purpose VLMs (GPT-5, OpenAI-o3, Qwen2.5-VL-7B) demonstrate stronger contextual awareness but remain less reliable than the proposed method. The Supervised Pairwise Classifier achieves higher accuracy than zero-shot VLMs, demonstrating that fine-tuning can help with aligning a model towards human preferences. DriveCritic outperforms all baselines by a significant margin, reaching 76.0\% accuracy, validating the effectiveness of the DriveCritic model and the proposed training paradigm.

\begin{table}[t]
\centering
\caption{Ablation on the DriveCritic training recipe.  
Checkmarks (\cmark) indicate enabled components. 
`Acc.' under Rewards denotes an accuracy-based reward. 
Final column reports accuracy on the DriveCritic test set.}
\label{tab:ablation_training}
\begin{tabular}{c c cc cc c}
\toprule
\multirow{2}{*}{\textbf{ID}} &
\multirow{2}{*}{\textbf{SFT}} &
\multicolumn{2}{c}{\textbf{RL}} &
\multicolumn{2}{c}{\textbf{Rewards}} &
\multirow{2}{*}{\textbf{Accuracy}} \\
\cmidrule(lr){3-4}\cmidrule(lr){5-6}
 &  & GRPO & DAPO & Format & Acc. &  \\
\midrule
A & \xmark & \xmark & \xmark & \xmark & \xmark & 0.480 \\
B & \xmark & \cellcolor{lightgreen}{\cmark} & \xmark & \cellcolor{lightgreen}{\cmark} & \cellcolor{lightgreen}{\cmark} & 0.464 \\
C & \cellcolor{lightgreen}{\cmark} & \xmark & \xmark & \xmark & \xmark & 0.645 \\
D & \cellcolor{lightgreen}{\cmark} & \cellcolor{lightgreen}{\cmark} & \xmark & \xmark & \cellcolor{lightgreen}{\cmark} & 0.739 \\
E & \cellcolor{lightgreen}{\cmark} & \cellcolor{lightgreen}{\cmark} & \xmark & \cellcolor{lightgreen}{\cmark} & \cellcolor{lightgreen}{\cmark} & 0.750 \\
F & \cellcolor{lightgreen}{\cmark} & \xmark & \cellcolor{lightgreen}{\cmark} & \cellcolor{lightgreen}{\cmark} & \cellcolor{lightgreen}{\cmark} & \textbf{0.760} \\
\bottomrule
\end{tabular}

\vspace{3pt}
\raggedright\footnotesize
\textbf{ID legend:}
A = base Qwen2.5-VL-7B (zero-shot);  
B = GRPO only (format + accuracy rewards);  
C = SFT only;  
D = SFT + GRPO (accuracy reward);  
E = SFT + GRPO (format + accuracy rewards);  
F = SFT + DAPO (format + accuracy rewards).
\vspace{-5mm}
\end{table}

\subsection{Qualitative Results}
\label{sec:qual}
In Fig.~\ref{fig:qualitative}, we show two qualitative examples where correct context understanding leads to aligning to the ground truth in the DriveCritic dataset. These examples show that fixed thresholds alone (e.g., lane offset, progress) can mis-rank trajectories in nuanced settings, while context-aware reasoning model (DriveCritic) can be used to complement rule-based evaluation methods in these challenging scenarios. We include more qualitative results in the supplementary video.

\subsection{Ablation Studies}
\label{sec:ablations}
We further conduct an ablation study to break down the contributions of components in the DriveCritic model. 
We note that only applying RL fine-tuning (B) could reduce the accuracy, highlighting the need of the SFT training (C) to warm up the model's ability. Building on SFT, all RL variants (D–F) yield clear gains, with the full DriveCritic recipe (F, SFT + DAPO~\cite{yu2025dapo} + format and accuracy rewards) achieving the best test accuracy on the DriveCritic dataset.

\begin{table}[t]
\centering
\caption{Robustness under trajectory-position flip on the DriveCritic test set. 
“No-flip acc.” and “flip acc.” are the standard accuracies before and after swapping the trajectory order. 
RR denotes robustness rate as defined above.}
\label{tab:robustness}
\begin{tabular}{lccc}
\toprule
\textbf{Model} & \textbf{No-flip acc.} & \textbf{Flip acc.} & \textbf{RR (\%)} \\
\midrule
Supervised Pairwise Classifier & 0.648 & 0.613 & 55.8 \\
Qwen2.5-VL-7B (base)          & 0.480 & 0.487 & 74.9 \\
Qwen2.5-VL-7B + SFT          & 0.645 & 0.649 & 78.0 \\
\textbf{DriveCritic (ours)}  & \textbf{0.760} & \textbf{0.765} & \textbf{81.8} \\
\bottomrule
\end{tabular}
\vspace{-7mm}
\end{table}

\subsection{Robustness Analysis}
\label{sec:robustness}
An important requirement for a learning-based driving evaluator is to produce \emph{consistent} judgments regardless of input ordering or formatting, a concern also raised in recent studies on LLM/VLM judges~\cite{ye2024justice,li2025generation}.  
To quantify robustness, we perform a \emph{position-flip test}: for every test pair, we swap the order of Trajectory~A and Trajectory~B in the prompt and re-evaluate the model.
Let $y^i$ be the original prediction and $\hat{y}^i$ the prediction after flipping. We follow~\cite{ye2024justice} to compute the \emph{Robustness Rate} (RR): $\mathrm{RR} = \frac{1}{|D|}\sum_{i=1}^{|D|} \mathbb{I}\bigl[y^i = \hat{y}^i\bigr]$, where $|D|$ is the size of the DriveCritic test set and $\mathbb{I}[\cdot]$ is the indicator function.
We also report the standard accuracy before (no-flip) and after (flip) swapping. As shown in Table~\ref{tab:robustness}, DriveCritic achieves the highest robustness rate (81.8\%), consistently outperforming other models. We observe that robustness improves steadily through the two-stage training pipeline, indicating that both SFT and RL fine-tuning contribute to stronger invariance to trajectory-order perturbations. Moreover, all VLM-based models maintain their accuracy after the flip, whereas the supervised classifier exhibits a notable drop, underscoring the advantage of a VLM backbone for this evaluation task.

\section{Limitations \& Outlook}
While DriveCritic improves alignment with human preferences, several limitations remain.
First, because it relies on a VLM, it inherits common weaknesses such as sensitivity to prompt design, domain shift, and occasional hallucinated or inconsistent judgments on out-of-distribution scenes. These issues are expected to diminish as stronger and more reliable VLMs emerge, and DriveCritic can directly benefit from such advances without architectural change.
Second, the curated preference dataset is still limited in scale and diversity, and our annotations currently rely on a single expert annotator. We did not measure inter-rater agreement, so label bias and limited generalizability remain concerns.
Third, DriveCritic does not currently leverage temporal information, which can lead to errors in dynamic scenarios such as changing traffic lights.
Finally, selective VLM-based adjudication introduces nontrivial computational and environmental costs. Although batching and caching help, the overhead can still be substantial at scale, posing practical and environmental challenges for large-scale deployment.

Despite these limitations, we see several promising directions for future work. 
Expanding the DriveCritic dataset across domains, evaluation modes, and driving styles~\cite{styledrive2025} will strengthen its utility.
Additionally, we think integrating the DriveCritic model to create a scalable human-aligned trajectory database with RL-based planners~\cite{li2025trajhf} is an interesting future direction to show that preference-aligned critics could guide RL fine-tuning of end-to-end planners.
Furthermore, exploring lighter-weight models or knowledge distillation from large VLMs to smaller student critics may reduce compute cost and improve deployability of VLM-based driving evaluation solutions.

\section{Conclusion}
In this work, we addressed the lack of context-awareness in state-of-the-art, rule-based metrics like EPDMS, which often misaligns with expert human judgment in complex driving scenarios. We introduced DriveCritic, a novel framework featuring a VLM evaluator and a new dataset of ambiguous scenarios annotated with pairwise human preferences. By fine-tuning the VLM with a two-stage supervised and reinforcement learning pipeline, our model learns to make human-aligned judgments. Our experiments validate this approach, showing DriveCritic achieves 76.0\% accuracy in aligning with human preferences, significantly outperforming all baselines. The model also demonstrates high robustness to input permutations, confirming the effectiveness of our training strategy. Ultimately, DriveCritic represents a significant step toward developing more reliable and human-centric evaluation tools for autonomous driving.

\noindent \textbf{GenAI Statement:} Generative AI tools were used solely for language editing.

\bibliographystyle{IEEEtran}
\bibliography{reference}

\begin{thebibliography}{10}
\providecommand{\url}[1]{#1}
\csname url@samestyle\endcsname
\providecommand{\newblock}{\relax}
\providecommand{\bibinfo}[2]{#2}
\providecommand{\BIBentrySTDinterwordspacing}{\spaceskip=0pt\relax}
\providecommand{\BIBentryALTinterwordstretchfactor}{4}
\providecommand{\BIBentryALTinterwordspacing}{\spaceskip=\fontdimen2\font plus
\BIBentryALTinterwordstretchfactor\fontdimen3\font minus \fontdimen4\font\relax}
\providecommand{\BIBforeignlanguage}[2]{{%
\expandafter\ifx\csname l@#1\endcsname\relax
\typeout{** WARNING: IEEEtran.bst: No hyphenation pattern has been}%
\typeout{** loaded for the language `#1'. Using the pattern for}%
\typeout{** the default language instead.}%
\else
\language=\csname l@#1\endcsname
\fi
#2}}
\providecommand{\BIBdecl}{\relax}
\BIBdecl

\bibitem{chen2024e2e_survey}
L.~Chen, P.~Wu, K.~Chitta, B.~Jaeger, A.~Geiger, and H.~Li, ``End-to-end autonomous driving: Challenges and frontiers,'' \emph{IEEE Transactions on Pattern Analysis and Machine Intelligence}, 2024.

\bibitem{mllm_driving_survey}
C.~Cui, Y.~Ma, X.~Cao, W.~Ye, Y.~Zhou, K.~Liang, J.~Chen, J.~Lu, Z.~Yang, K.-D. Liao, T.~Gao, E.~Li, K.~Tang, Z.~Cao, T.~Zhou, A.~Liu, X.~Yan, S.~Mei, J.~Cao, Z.~Wang, and C.~Zheng, ``A survey on multimodal large language models for autonomous driving,'' in \emph{Proceedings of the IEEE/CVF Winter Conference on Applications of Computer Vision (WACV) Workshops}, January 2024, pp. 958--979.

\bibitem{caesar2021nuplan}
H.~Caesar, J.~Kabzan, K.~S. Tan, W.~K. Fong, E.~Wolff, A.~Lang, L.~Fletcher, O.~Beijbom, and S.~Omari, ``nuplan: A closed-loop ml-based planning benchmark for autonomous vehicles,'' \emph{arXiv preprint arXiv:2106.11810}, 2021.

\bibitem{li2024hydramdp}
Z.~Li, K.~Li, S.~Wang, S.~Lan, Z.~Yu, Y.~Ji, Z.~Li, Z.~Zhu, J.~Kautz, Z.~Wu \emph{et~al.}, ``Hydra-mdp: End-to-end multimodal planning with multi-target hydra-distillation,'' \emph{arXiv preprint arXiv:2406.06978}, 2024.

\bibitem{li2025hydranext}
Z.~Li, S.~Wang, S.~Lan, Z.~Yu, Z.~Wu, and J.~M. Alvarez, ``Hydra-next: Robust closed-loop driving with open-loop training,'' \emph{arXiv preprint arXiv:2503.12030}, 2025.

\bibitem{chen2024vadv2}
S.~Chen, B.~Jiang, H.~Gao, B.~Liao, Q.~Xu, Q.~Zhang, C.~Huang, W.~Liu, and X.~Wang, ``Vadv2: End-to-end vectorized autonomous driving via probabilistic planning,'' \emph{arXiv preprint arXiv:2402.13243}, 2024.

\bibitem{caesar2020nuscenes}
H.~Caesar, V.~Bankiti, A.~H. Lang, S.~Vora, V.~E. Liong, Q.~Xu, A.~Krishnan, Y.~Pan, G.~Baldan, and O.~Beijbom, ``nuscenes: A multimodal dataset for autonomous driving,'' in \emph{Proceedings of the IEEE/CVF conference on computer vision and pattern recognition}, 2020, pp. 11\,621--11\,631.

\bibitem{Dauner2024NEURIPS}
D.~Dauner, M.~Hallgarten, T.~Li, X.~Weng, Z.~Huang, Z.~Yang, H.~Li, I.~Gilitschenski, B.~Ivanovic, M.~Pavone, A.~Geiger, and K.~Chitta, ``Navsim: Data-driven non-reactive autonomous vehicle simulation and benchmarking,'' in \emph{Advances in Neural Information Processing Systems (NeurIPS)}, 2024.

\bibitem{zhai2023ad-mlp}
J.-T. Zhai, Z.~Feng, J.~Du, Y.~Mao, J.-J. Liu, Z.~Tan, Y.~Zhang, X.~Ye, and J.~Wang, ``Rethinking the open-loop evaluation of end-to-end autonomous driving in nuscenes,'' \emph{arXiv preprint arXiv:2305.10430}, 2023.

\bibitem{dosovitskiy2017carla}
A.~Dosovitskiy, G.~Ros, F.~Codevilla, A.~Lopez, and V.~Koltun, ``Carla: An open urban driving simulator,'' in \emph{CoRL}.\hskip 1em plus 0.5em minus 0.4em\relax PMLR, 2017, pp. 1--16.

\bibitem{waymo_womd_2021}
S.~Ettinger, S.~Cheng, B.~Caine, C.~Liu, H.~Zhao, S.~Pradhan, Y.~Chai, B.~Sapp, C.~Qi, Y.~Zhou \emph{et~al.}, ``Large scale interactive motion forecasting for autonomous driving: The waymo open motion dataset,'' in \emph{ICCV}, 2021.

\bibitem{sun2020waymo_open_dataset}
P.~Sun, H.~Kretzschmar, X.~Dotiwalla, A.~Chouard, V.~Patnaik, P.~Tsui, J.~Guo, Y.~Zhou, Y.~Chai, B.~Caine \emph{et~al.}, ``Scalability in perception for autonomous driving: Waymo open dataset,'' in \emph{Proceedings of the IEEE/CVF conference on computer vision and pattern recognition}, 2020, pp. 2446--2454.

\bibitem{Cao2025CORL}
W.~Cao, M.~Hallgarten, T.~Li, D.~Dauner, X.~Gu, C.~Wang, Y.~Miron, M.~Aiello, H.~Li, I.~Gilitschenski, B.~Ivanovic, M.~Pavone, A.~Geiger, and K.~Chitta, ``Pseudo-simulation for autonomous driving,'' in \emph{CoRL}, 2025.

\bibitem{waymo2025E2Echallenge}
``2025 vision-based end-to-end driving challenge,'' \url{https://waymo.com/open/challenges/2025/e2e-driving/}, 2025.

\bibitem{li2025gtrs}
Z.~Li, W.~Yao, Z.~Wang, X.~Sun, J.~Chen, N.~Chang, M.~Shen, Z.~Wu, S.~Lan, and J.~M. Alvarez, ``Generalized trajectory scoring for end-to-end multimodal planning,'' \emph{arXiv preprint arXiv:2506.06664}, 2025.

\bibitem{styledrive2025}
R.~Hao, B.~Jing, H.~Yu, and Z.~Nie, ``Styledrive: Towards driving-style aware benchmarking of end-to-end autonomous driving,'' \emph{arXiv preprint arXiv:2506.23982}, 2025.

\bibitem{li2025recogdrive}
Y.~Li, K.~Xiong, X.~Guo, F.~Li, S.~Yan, G.~Xu, L.~Zhou, L.~Chen, H.~Sun, B.~Wang \emph{et~al.}, ``Recogdrive: A reinforced cognitive framework for end-to-end autonomous driving,'' \emph{arXiv preprint arXiv:2506.08052}, 2025.

\bibitem{achiam2023gpt4_report}
J.~Achiam, S.~Adler, S.~Agarwal, L.~Ahmad, I.~Akkaya, F.~L. Aleman, D.~Almeida, J.~Altenschmidt, S.~Altman, S.~Anadkat \emph{et~al.}, ``Gpt-4 technical report,'' \emph{arXiv preprint arXiv:2303.08774}, 2023.

\bibitem{openai2025gpt5}
{OpenAI}, ``Gpt-5,'' \url{https://openai.com/gpt-5/}, 2025.

\bibitem{openai2025o3}
OpenAI, ``Introducing openai o3 and o4-mini,'' \url{https://openai.com/index/introducing-o3-and-o4-mini/}, 2025.

\bibitem{team2023gemini}
G.~Team, R.~Anil, S.~Borgeaud, J.-B. Alayrac, J.~Yu, R.~Soricut, J.~Schalkwyk, A.~M. Dai, A.~Hauth, K.~Millican \emph{et~al.}, ``Gemini: a family of highly capable multimodal models,'' \emph{arXiv preprint arXiv:2312.11805}, 2023.

\bibitem{Qwen2.5-VL}
S.~Bai, K.~Chen, X.~Liu, J.~Wang, W.~Ge, S.~Song, K.~Dang, P.~Wang, S.~Wang, J.~Tang, H.~Zhong, Y.~Zhu, M.~Yang, Z.~Li, J.~Wan, P.~Wang, W.~Ding, Z.~Fu, Y.~Xu, J.~Ye, X.~Zhang, T.~Xie, Z.~Cheng, H.~Zhang, Z.~Yang, H.~Xu, and J.~Lin, ``Qwen2.5-vl technical report,'' \emph{arXiv preprint arXiv:2502.13923}, 2025.

\bibitem{shao2024deepseekmath}
Z.~Shao, P.~Wang, Q.~Zhu, R.~Xu, J.~Song, X.~Bi, H.~Zhang, M.~Zhang, Y.~Li, Y.~Wu \emph{et~al.}, ``Deepseekmath: Pushing the limits of mathematical reasoning in open language models,'' \emph{arXiv preprint arXiv:2402.03300}, 2024.

\bibitem{yu2025dapo}
Q.~Yu, Z.~Zhang, R.~Zhu, Y.~Yuan, X.~Zuo, Y.~Yue, W.~Dai, T.~Fan, G.~Liu, L.~Liu \emph{et~al.}, ``Dapo: An open-source llm reinforcement learning system at scale,'' \emph{arXiv preprint arXiv:2503.14476}, 2025.

\bibitem{yue2025rlvr}
Y.~Yue, Z.~Chen, R.~Lu, A.~Zhao, Z.~Wang, S.~Song, and G.~Huang, ``Does reinforcement learning really incentivize reasoning capacity in llms beyond the base model?'' \emph{arXiv preprint arXiv:2504.13837}, 2025.

\bibitem{jia2024bench2drive}
X.~Jia, Z.~Yang, Q.~Li, Z.~Zhang, and J.~Yan, ``Bench2drive: Towards multi-ability benchmarking of closed-loop end-to-end autonomous driving,'' \emph{NeurIPS}, vol.~37, pp. 819--844, 2024.

\bibitem{hu2023uniad}
Y.~Hu, J.~Yang, L.~Chen, K.~Li, C.~Sima, X.~Zhu, S.~Chai, S.~Du, T.~Lin, W.~Wang \emph{et~al.}, ``Planning-oriented autonomous driving,'' in \emph{Proceedings of the IEEE/CVF conference on computer vision and pattern recognition}, 2023, pp. 17\,853--17\,862.

\bibitem{ego_status_all_you_need}
Z.~Li, Z.~Yu, S.~Lan, J.~Li, J.~Kautz, T.~Lu, and J.~M. Alvarez, ``Is ego status all you need for open-loop end-to-end autonomous driving?'' in \emph{Proceedings of the IEEE/CVF Conference on Computer Vision and Pattern Recognition (CVPR)}, June 2024, pp. 14\,864--14\,873.

\bibitem{mu2024most}
N.~Mu, J.~Ji, Z.~Yang, N.~Harada, H.~Tang, K.~Chen, C.~R. Qi, R.~Ge, K.~Goel, Z.~Yang \emph{et~al.}, ``Most: Multi-modality scene tokenization for motion prediction,'' in \emph{CVPR}, 2024, pp. 14\,988--14\,999.

\bibitem{tiandrivevlm}
X.~Tian, J.~Gu, B.~Li, Y.~Liu, Y.~Wang, Z.~Zhao, K.~Zhan, P.~Jia, X.~Lang, and H.~Zhao, ``Drivevlm: The convergence of autonomous driving and large vision-language models,'' in \emph{8th Annual Conference on Robot Learning}.

\bibitem{sima2024drivelm}
C.~Sima, K.~Renz, K.~Chitta, L.~Chen, H.~Zhang, C.~Xie, J.~Bei{\ss}wenger, P.~Luo, A.~Geiger, and H.~Li, ``Drivelm: Driving with graph visual question answering,'' in \emph{European conference on computer vision}.\hskip 1em plus 0.5em minus 0.4em\relax Springer, 2024, pp. 256--274.

\bibitem{wang2025omnidrive}
S.~Wang, Z.~Yu, X.~Jiang, S.~Lan, M.~Shi, N.~Chang, J.~Kautz, Y.~Li, and J.~M. Alvarez, ``Omnidrive: A holistic vision-language dataset for autonomous driving with counterfactual reasoning,'' in \emph{Proceedings of the Computer Vision and Pattern Recognition Conference}, 2025, pp. 22\,442--22\,452.

\bibitem{jiang2024senna}
B.~Jiang, S.~Chen, B.~Liao, X.~Zhang, W.~Yin, Q.~Zhang, C.~Huang, W.~Liu, and X.~Wang, ``Senna: Bridging large vision-language models and end-to-end autonomous driving,'' \emph{arXiv preprint arXiv:2410.22313}, 2024.

\bibitem{jiang2025alphadrive}
B.~Jiang, S.~Chen, Q.~Zhang, W.~Liu, and X.~Wang, ``Alphadrive: Unleashing the power of vlms in autonomous driving via reinforcement learning and reasoning,'' \emph{arXiv preprint arXiv:2503.07608}, 2025.

\bibitem{qian2025agentthink}
K.~Qian, S.~Jiang, Y.~Zhong, Z.~Luo, Z.~Huang, T.~Zhu, K.~Jiang, M.~Yang, Z.~Fu, J.~Miao \emph{et~al.}, ``Agentthink: A unified framework for tool-augmented chain-of-thought reasoning in vision-language models for autonomous driving,'' \emph{arXiv preprint arXiv:2505.15298}, 2025.

\bibitem{ye2024justice}
J.~Ye, Y.~Wang, Y.~Huang, D.~Chen, Q.~Zhang, N.~Moniz, T.~Gao, W.~Geyer, C.~Huang, P.-Y. Chen \emph{et~al.}, ``Justice or prejudice? quantifying biases in llm-as-a-judge,'' \emph{arXiv preprint arXiv:2410.02736}, 2024.

\bibitem{li2025generation}
D.~Li, B.~Jiang, L.~Huang, A.~Beigi, C.~Zhao, Z.~Tan, A.~Bhattacharjee, Y.~Jiang, C.~Chen, T.~Wu \emph{et~al.}, ``From generation to judgment: Opportunities and challenges of llm-as-a-judge, 2025,'' \emph{URL https://arxiv. org/abs/2411.16594}, 2025.

\bibitem{liualigning}
Y.~Liu, H.~Zhou, Z.~Guo, E.~Shareghi, I.~Vuli{\'c}, A.~Korhonen, and N.~Collier, ``Aligning with human judgement: The role of pairwise preference in large language model evaluators,'' in \emph{First Conference on Language Modeling}.

\bibitem{you2025comprehensive}
S.~You, X.~Luo, X.~Liang, J.~Yu, C.~Zheng, and J.~Gong, ``A comprehensive llm-powered framework for driving intelligence evaluation,'' \emph{arXiv preprint arXiv:2503.05164}, 2025.

\bibitem{xie2025drivebench}
S.~Xie, L.~Kong, Y.~Dong, C.~Sima, W.~Zhang, Q.~A. Chen, Z.~Liu, and L.~Pan, ``Are vlms ready for autonomous driving? an empirical study from the reliability, data, and metric perspectives,'' \emph{arXiv preprint arXiv:2501.04003}, 2025.

\bibitem{wang2024he}
J.~Wang, X.~Zhang, Z.~Xing, S.~Gu, X.~Guo, Y.~Hu, Z.~Song, Q.~Zhang, X.~Long, and W.~Yin, ``He-drive: Human-like end-to-end driving with vision language models,'' \emph{arXiv preprint arXiv:2410.05051}, 2024.

\bibitem{li2025trajhf}
D.~Li, J.~Ren, Y.~Wang, X.~Wen, P.~Li, L.~Xu, K.~Zhan, Z.~Xia, P.~Jia, X.~Lang, N.~Xu, and H.~Zhao, ``Finetuning generative trajectory model with reinforcement learning from human feedback,'' \emph{arXiv preprint arXiv:2503.10434}, 2025.

\bibitem{contributors2023openscene}
O.~Contributors, ``Openscene: The largest up-to-date 3d occupancy prediction benchmark in autonomous driving,'' in \emph{Proceedings of the Conference on Computer Vision and Pattern Recognition, Vancouver, Canada}, 2023, pp. 18--22.

\bibitem{qin2024large}
Z.~Qin, R.~Jagerman, K.~Hui, H.~Zhuang, J.~Wu, L.~Yan, J.~Shen, T.~Liu, J.~Liu, D.~Metzler \emph{et~al.}, ``Large language models are effective text rankers with pairwise ranking prompting,'' in \emph{Findings of the Association for Computational Linguistics: NAACL 2024}, 2024, pp. 1504--1518.

\bibitem{zheng2024llamafactory}
\BIBentryALTinterwordspacing
Y.~Zheng, R.~Zhang, J.~Zhang, Y.~Ye, Z.~Luo, Z.~Feng, and Y.~Ma, ``Llamafactory: Unified efficient fine-tuning of 100+ language models,'' in \emph{Proceedings of the 62nd Annual Meeting of the Association for Computational Linguistics (Volume 3: System Demonstrations)}.\hskip 1em plus 0.5em minus 0.4em\relax Bangkok, Thailand: Association for Computational Linguistics, 2024. [Online]. Available: \url{http://arxiv.org/abs/2403.13372}
\BIBentrySTDinterwordspacing

\bibitem{zheng2025easyr1}
Y.~Zheng, J.~Lu, S.~Wang, Z.~Feng, D.~Kuang, and Y.~Xiong, ``Easyr1: An efficient, scalable, multi-modality rl training framework,'' \url{https://github.com/hiyouga/EasyR1}, 2025.

\bibitem{sheng2024verl}
G.~Sheng, C.~Zhang, Z.~Ye, X.~Wu, W.~Zhang, R.~Zhang, Y.~Peng, H.~Lin, and C.~Wu, ``Hybridflow: A flexible and efficient rlhf framework,'' \emph{arXiv preprint arXiv: 2409.19256}, 2024.

\bibitem{he2016resnet}
K.~He, X.~Zhang, S.~Ren, and J.~Sun, ``Deep residual learning for image recognition,'' in \emph{Proceedings of the IEEE conference on computer vision and pattern recognition}, 2016, pp. 770--778.

\end{thebibliography}

\end{document}